\PassOptionsToPackage{pdftex,dvipsnames}{xcolor}
\documentclass[letterpaper, 10 pt, conference]{ieeeconf}  

\IEEEoverridecommandlockouts                             
\usepackage{amsmath}
\usepackage{amssymb}
\let\labelindent\relax
\usepackage{enumitem}

\usepackage{amsmath,amsfonts}
\usepackage[font=footnotesize]{caption}
\usepackage{subcaption}
\usepackage{algorithmic}
\usepackage{algorithm}
\usepackage{array}
\usepackage{textcomp}
\usepackage{url}
\usepackage{graphicx}
\usepackage{tabularray}
\usepackage{float}
\usepackage{bm}
\usepackage{mathtools}
\usepackage{stfloats}
\usepackage{mathrsfs}
\usepackage{multirow}
\usepackage{makecell}
\usepackage{vcell}
\usepackage{booktabs}
\usepackage{diagbox}
\usepackage{csquotes}
\usepackage{tabularx}
\usepackage{tikz}
\usepackage{lipsum}
\usepackage{url}
\usepackage{schemata}
\usepackage{verbatim}
\usepackage{pifont}

\usepackage{cite}
\makeatletter
\let\NAT@parse\undefined
\makeatother
\usepackage[hidelinks]{hyperref}
\hypersetup{
colorlinks=true,
linkcolor=blue,
filecolor=magenta,
urlcolor=blue,
}
\hypersetup{breaklinks=true}
\begin{document}

\title{\LARGE \bf
Few-Shot Demonstration-Driven Task Coordination and Trajectory Execution for Multi-Robot Systems
}

\author{Taehyeon Kim$^{1}\dagger$, Vishnunandan L. N. Venkatesh$^{2}\dagger$, and Byung-Cheol Min$^{3}$
\thanks{$^{1}$Purdue University, West Lafayette, IN, USA {\tt\footnotesize kim4435@purdue.edu}.}%
\thanks{$^{2}$Astemo Americas, Inc., Farmington Hills, MI, USA {\tt\footnotesize vishnunandan.venkatesh.on@astemo.com}.}%
\thanks{$^{3}$Indiana University Bloomington, Bloomington, IN, USA {\tt\footnotesize minb@iu.edu}.}%
\thanks{$\dagger$ Equal contribution.}%
}

\maketitle

\begin{abstract}


Learning coordinated behaviors for multi-robot systems from only a few demonstrations is difficult because temporal task dependencies and spatial trajectory generation are tightly coupled, which increases the hypothesis space and often yields unstable generalization in data-scarce regimes. We present DDACE, a structured few-shot learning framework that introduces a structural inductive bias by explicitly decoupling temporal coordination from spatial trajectory synthesis. Demonstrations are first processed via spectral clustering to extract coordination structure and form interaction graphs. A Temporal Graph Network predicts action dependencies and sequences, while Gaussian Process models generate progress-parameterized geometric trajectories that adapt to new start/goal configurations. This factorized design reduces hypothesis coupling and improves data efficiency for few-shot multi-robot coordination. Extensive simulation studies and real-robot experiments show that DDACE produces stable coordinated executions from a small number of demonstrations and improves trajectory consistency compared to end-to-end imitation baselines under limited data. Additional materials are available at \url{https://sites.google.com/view/ddace}.

\end{abstract}

\section{Introduction}
Learning from Demonstration (LfD) has emerged as a transformative paradigm in robotics, enabling systems to acquire complex behaviors efficiently by leveraging human-provided demonstrations. Bypassing the need for explicitly programmed rules and reward engineering, LfD allows robots to generalize skills from demonstrations, significantly lowering the barrier to the deployment of intelligent systems in various domains \cite{ravichandar2020recent, argall2009survey}. The applicability of LfD spans a wide range of fields, from industrial automation to service robotics, where human expertise can be intuitively transferred to robotic agents~\cite{billard2016learning}.

Although effective in single-agent settings, applying LfD to Multi-Robot Systems (MRS) introduces unique challenges. Multi-robot tasks often involve intricate spatial–temporal interdependencies\cite{burgard2005coordinated}, requiring synchronized behaviors across heterogeneous agents with diverse capabilities \cite{darmanin2017review, knepper2013ikeabot}. This is critical in sequential task execution (e.g., sports, collaborative assembly) and heterogeneous team coordination (e.g., logistics, disaster response), where learning temporal action sequences - rather than only final goal states - is essential for robust coordination.

\begin{figure}[!t]
\centering
\includegraphics[width=0.45\textwidth]{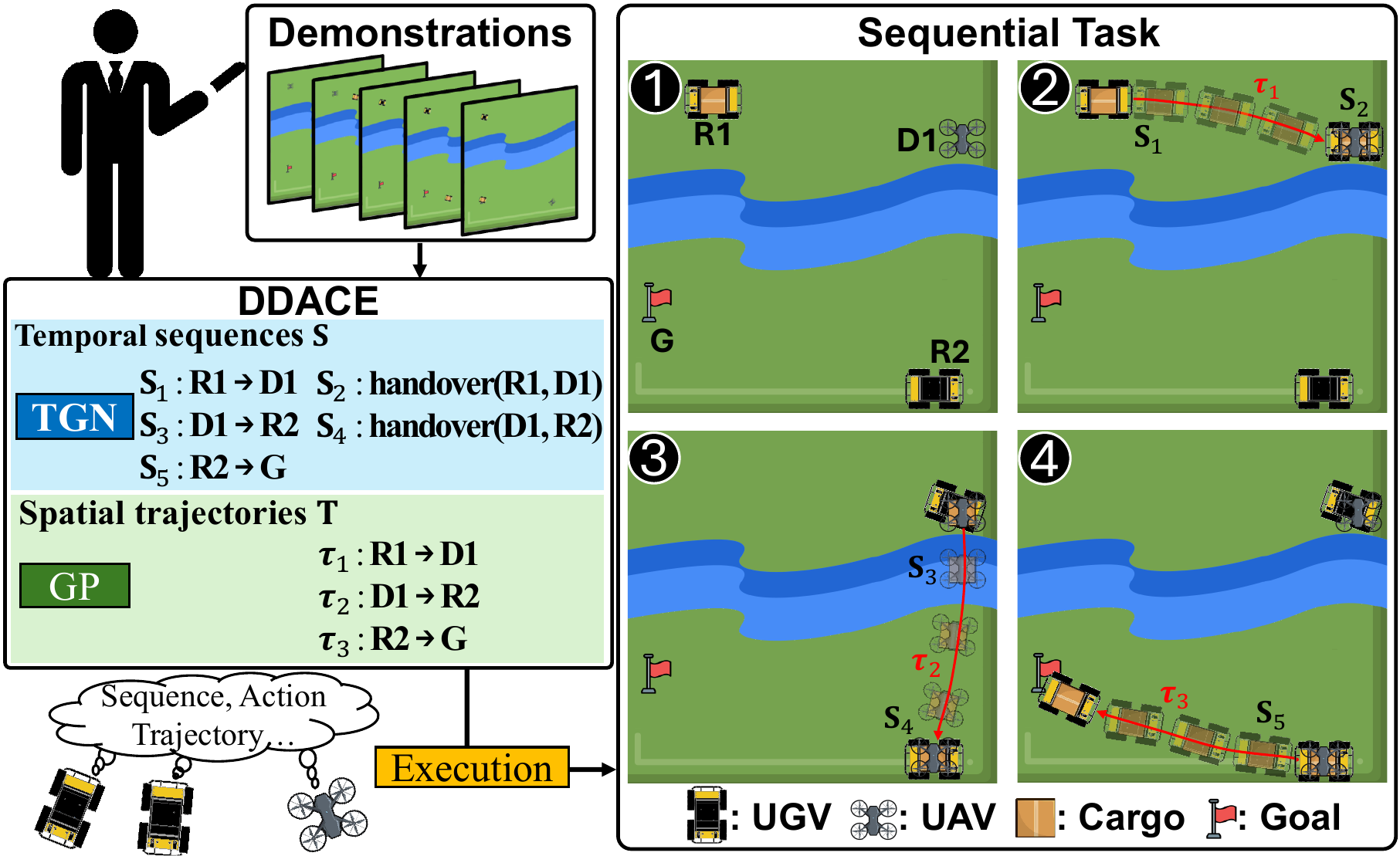}
\caption{Concept for few-shot Demonstration-Driven tAsk Coordination and trajectory Execution for multi-robot systems (DDACE). A human expert demonstrates the task, from which the robots learn both sequences, composed of multiple actions and trajectories.}
\label{fig_1}
\vspace{-10pt}
\end{figure}

However, most LfD methods are data-hungry, task-specific, and overlook modeling of spatial and temporal aspects, limiting generalization in few-shot settings \cite{argall2009survey, balakuntala2021learning}. Many focus solely on achieving goal states, neglecting action sequencing, which is vital in applications demanding trajectory fidelity. Although some domain-specific work, such as robot soccer \cite{freelan2015towards, simoes2020dataset}, considers sequential behaviors, it usually relies on specialized reward functions, constraining adaptability. Consequently, there remains a critical need for an LfD framework that not only operates efficiently under few-shot constraints but also generalizes across multi-robot tasks while explicitly modeling the temporal dependencies crucial for effective coordination.

End-to-end architectures typically model temporal coordination and spatial motion generation jointly within a single learning model. Here, temporal coordination refers to action ordering and inter-robot dependencies, while spatial motion generation refers to the geometric trajectories used to execute those actions. Although such joint modeling can be effective with sufficient data, in few-shot regimes it increases hypothesis coupling across representational domains, enlarges effective capacity, and heightens overfitting risk, which can destabilize multi-robot coordination. DDACE instead adopts temporal-spatial decomposition, combined with spectral relational sparsification, as a structural inductive bias to reduce hypothesis complexity and stabilize coordination under limited demonstrations. Under this formulation, DDACE factorizes sequence learning and trajectory generation: a Temporal Graph Network (TGN) constructed over spectrally refined relational graphs models high-level action dependencies, while Gaussian Processes (GPs) synthesize geometry-consistent motion primitives. The framework is evaluated on controlled multi-robot task families designed to analyze coordination under data scarcity.

The key contributions of this work are summarized as follows:
\begin{itemize}[leftmargin=*]

\item A structural formulation for few-shot multi-robot learning is introduced, in which temporal-spatial decomposition is coupled with relational sparsification as an explicit inductive bias, reducing hypothesis coupling across representational domains and improving stability under limited demonstrations.

\item The DDACE framework is presented as a modular instantiation of this inductive bias for multi-robot coordination, integrating a spectrally refined TGN for action-sequence and partial-order inference with GP motion models for progress-parameterized, geometry-consistent trajectory synthesis and adaptation to novel start-goal configurations.

\item Controlled structural studies are conducted across multi-robot task families that vary sequence length, team size, action-space complexity, and trajectory geometry, enabling analysis of how decomposition and sparsity affect stability and generalization in few-shot regimes.
\end{itemize}

\section{Related Works}
LfD enables multi-robot systems to acquire complex behaviors through intuitive, human-led instruction. Methods include teleoperation-based learning for navigation and manipulation \cite{martins2010learning,sullivan2011hierarchical}, reinforcement learning with state-space automata for robot soccer \cite{freelan2014towards,simoes2020dataset}, and adaptive confidence-based mechanisms for interactive demonstrations \cite{chernova2009interactive}. Others use geometric CAD models for structured collaboration \cite{knepper2013ikeabot} or vision-based demonstrations with probabilistic models such as Gaussian Mixture Models for bi-manual tasks \cite{huang2017vision}. These approaches often face challenges of task specificity, reliance on large datasets, and limited intuitive, vision-based communication \cite{ravichandar2020recent,argall2009survey,brunke2022safe}. More recent frameworks integrate interaction keypoints and reinforcement learning for coordination \cite{venkatesh2024learning}, but do not explicitly address spatiotemporal modeling.

Graph Neural Networks (GNNs) naturally represent MRS as graphs, with robots as nodes and interactions as edges \cite{blumenkamp2022framework}, capturing relational dependencies essential for coordination \cite{chandrasekaran2023survey}. Applications include decentralized path planning for collision-free navigation in partially observable environments \cite{li2020graph, chandrasekaran2023survey}, with prior work also demonstrating decentralized GNN policies on real multi-robot systems \cite{blumenkamp2022framework}. In task allocation, GNNs solve combinatorial problems such as the Linear Sum Assignment Problem (LSAP) using heterogeneous graphs \cite{wu2024multi, goarin2024graph}, sometimes enhanced with heuristics like the Consensus-Based Bundle Algorithm (CBBA) \cite{chekakta2024towards}. They also enable cooperative behaviors-flocking, coverage, and formation control \cite{blumenkamp2022framework, zhang2022h2gnn, chen2021spatio, chekakta2024towards, goeckner2024graph}. Hierarchical variants like H2GNN further improve exploration efficiency by aggregating spatial information \cite{zhang2022h2gnn}.

Recent state-of-the-art end-to-end multi-agent learning approaches, such as MADiff \cite{zhu2024madiff} and MADT \cite{meng2023offline}, demonstrate strong coordination in offline MARL domains. MADiff uses diffusion models to jointly generate multi-agent action or trajectory sequences, while MADT applies transformer-based sequence modeling for decision making. Both approaches rely on large offline datasets, which is incompatible with few-shot demonstration learning, and integrate temporal and spatial learning within a single model, limiting modularity and adaptability. They also omit progress parameterized geometry for flexible start and goal adaptation-capabilities critical for few-shot multi-robot learning.

In contrast, DDACE employs a hierarchically decoupled framework that learns temporal action sequencing and spatial motion generation independently. It generalizes from minimal demonstrations by extracting essential robot–object interactions without large datasets or handcrafted rewards, unlike GNN-based or end-to-end MRS approaches, it leverages graph-based temporal reasoning and progress-parameterized geometry for rapid, few-shot adaptation to diverse multi-robot coordination tasks.

\section{Methodology}

\subsection{Problem Definition}

Few-shot multi-robot coordination requires learning both temporal dependencies and spatial motion patterns from limited demonstrations. When these components are jointly modeled, hypothesis coupling increases across representational domains, enlarging effective capacity and reducing learning stability under data scarcity. To mitigate this structural challenge, a factorized formulation is adopted in which temporal reasoning and spatial trajectory generation are modeled independently. This decomposition, combined with relational sparsification, serves as a structural inductive bias that constrains hypothesis complexity.

Under this formulation, the learning objective is defined over two decoupled components: temporal sequence inference and trajectory synthesis, as described below.

Let $D = \{d_1, d_2, \dots, d_k\}$ be a set of demonstrations for coordinated multi-robot tasks.  
Each demonstration $d_i$ is represented by a \emph{demonstration graph} $G_i = (N_i, E_i)$, where:  
\begin{itemize}[leftmargin=*]
    \item $N_i$ represents robots, goals, and other interactive objects.
    \item $E_i$ encodes \emph{interdependencies} such as enabling relations (finish~$\!\to\!$~start) or physical interaction constraints.
\end{itemize}

Each $d_i$ contains two data modalities, reflecting the structure of a hierarchical policy:
\begin{itemize}[leftmargin=*]
\item Temporal sequences $\mathbf{S}^{\text{demo}}_i$: ordered action tokens specifying \emph{when} each robot-specific operation occurs, capturing high-level task logic and partial-order constraints without prescribing motion geometry.
\item Spatial trajectories $\mathbf{T}^{\text{demo}}_i$: continuous paths from $x_0$ to $x_T$, reparameterized by progress $s \in [0,1]$ so that geometry is modeled independently of absolute time.
\end{itemize}

The \emph{spatial} modality is a geometry-only path $\mathbf{x}(s)$, $s\in[0,1]$, and the \emph{temporal} modality is a partial order $\langle\mathcal{A},\prec\rangle$ over action tokens, with $\prec$ from inferred enabling edges; only ordering is modeled, not durations.

Demonstrations are assumed to be optimal and not noisy, encoding coherent multi-step task sequences that can be used to infer both temporal partial orders and spatial trajectories. This assumption ensures that meaningful patterns can be extracted from limited data in few-shot learning settings.

The learning objective is defined as: $F: D \rightarrow (\mathbf{S}, \mathbf{T})$,
where $\mathbf{S}$ constitutes the \textit{high-level policy} (temporal coordination) and $\mathbf{T}$ constitutes the \textit{low-level policy} (spatial motion generation), with temporal reasoning explicitly decoupled from spatial processing.

\subsubsection{Sequence Learning via Temporal Graph Networks} 
\begin{equation}
f_T\left(C(G_i); \theta\right) \approx \mathbf{S}^{\text{demo}}_i
\end{equation}
Here $C(G_i)$ denotes a spectral clustering of $G_i$ to extract key inter-node relations.  
A Temporal Graph Network (TGN), parameterized by $\theta$, is used to learn the temporal ordering such that predicted sequences match those in the demonstrations.

\vspace{4pt}
\subsubsection{Trajectory Learning via Gaussian Processes}
\begin{equation}
f_S\left(x_0, x_T; \phi\right) \approx \mathbf{T}^{\text{demo}}_i
\end{equation}
Here $\phi$ denotes the Gaussian Process hyperparameters.  
With z-transform scaling in a canonical coordinate frame, $f_S$ reproduces demonstrated paths and synthesizes novel, geometrically consistent trajectories without committing to specific execution timing.

\vspace{4pt}
\subsubsection{Execution Semantics}
At decision epochs (periodic or event-triggered), the high-level policy $f_T$ selects the next action consistent with the learned partial order.
For motion tokens $a \in \mathcal{A}$, the low-level policy $f_S$ outputs a progress-parameterized path $\mathbf{x}_a(s)$, tracked by a controller.
This separation ensures temporal logic governs \emph{when} actions occur and spatial processing governs \emph{how} motion is executed, avoiding ambiguity between geometry and timestamps.

The objective is to learn $f_T$ and $f_S$ such that their outputs jointly reproduce and generalize $\mathbf{S}^{\text{demo}}_i$ and $\mathbf{T}^{\text{demo}}_i$ for few-shot coordinated multi-robot tasks.

\begin{figure*}
\centering
\includegraphics[width=0.90\textwidth]{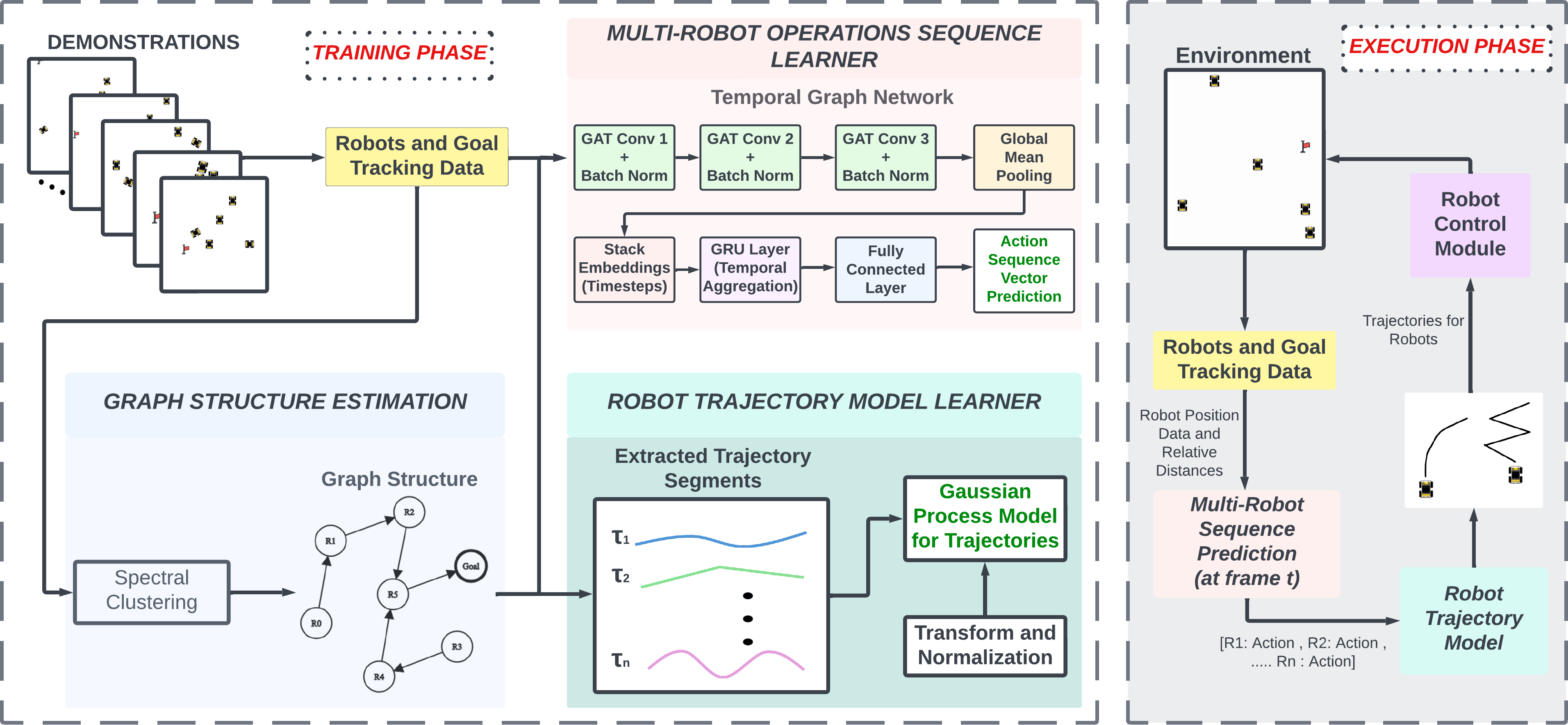}
\vspace{-2pt}
\caption{Overview of the proposed DDACE framework. In the training phase, demonstration data is preprocessed, graph structures are extracted via spectral clustering, and the Temporal Graph Networks (TGN) and Gaussian Processes (GP) models are trained independently. During the execution phase, the trained models predict coordinated action sequences and generate spatial trajectories for new scenarios, enabling adaptive and efficient multi-robot task execution.}
\label{fig_2}
\vspace{-10pt}
\end{figure*}

\subsection{Temporal Graph Networks (TGNs)}
\subsubsection{Preprocessing}
Let $d_i \in D$ be a multi-robot task demonstration recorded as a CSV, where each row corresponds to a discrete time step $t \in {1,\dots, T_i}$ and contains node features and labels. Node features are the $(x,y)$ coordinates of each entity in $N_i$ (e.g., robots, goals, objects), while labels indicate their action state or activity. From each $d_i$, node features are reshaped into $\mathbf{X}_i \in \mathbb{R}^{T_i \times |N_i| \times 2}$ and labels into $\mathbf{L}_i \in \mathbb{R}^{T_i \times |N_i|}$, capturing spatial positions and dynamic state changes over time.

A critical component of our preprocessing is the automatic segmentation of demonstrations into high-level action steps. This is achieved by selecting keyframes from $\mathbf{L}_i$ where a nonzero activity label appears and differs from the previous frame:
\begin{equation}
\mathbf{L}_i(t) \neq \mathbf{0} \quad \text{and} \quad \mathbf{L}_i(t) \neq \mathbf{L}_i(t-1).
\end{equation}
This rule infers substep boundaries directly from observed state changes, ensuring they are fully derived from demonstration data without reliance on manually pre-segmented inputs.

The final output is a sequence of graph snapshots containing node features and labels, condensed to key events to reduce redundancy and focus learning on critical transitions for robust multi-robot coordination and manipulation.

\subsubsection{Spectral Clustering}

From the keyframe sequence of $d_i$, directed edges $E_i$ are extracted and aggregated across demonstrations into the unique edge set $\mathcal{E}=\{e_1,\dots,e_m\}$. Edge frequencies are computed as
\begin{equation}
F(e) = \sum_{i=1}^{k} f_{e}(d_i),
\end{equation}
capturing persistent inter-node interactions.

A graph Laplacian $\mathbf{L} = \mathbf{D} - \mathbf{A}$ is constructed from the weighted adjacency structure, and spectral clustering is applied to retain structurally consistent relational patterns. This produces a refined edge set
\begin{equation}
E' = \{e \in \mathcal{E} \mid \mathcal{C}(e) \ \text{is assigned}\},
\end{equation}
which guides spatial aggregation in the TGN.

By preserving only persistent relational dependencies, this refinement injects structural sparsity and improves robustness in few-shot multi-robot coordination.





\subsubsection{Graph Attention Networks and Gated Recurrent Units}

Spatial and temporal dependencies are modeled in a decoupled manner. Spatial encoding captures relational configurations at each keyframe, while temporal encoding models their evolution across keyframes.

Let $\mathbf{X}_i^{(t)} \in \mathbb{R}^{|N_i| \times 2}$ denote the node feature matrix at keyframe $t$ of demonstration $d_i \in D$, where the rows correspond to entities in the node set $N_i$. In the spatial encoding stage, these features are processed through a series of GATs. Importantly, the connectivity structure, represented by the refined edge index $\mathbf{E}'$, is derived from the spectral clustering procedure described above.

We initialize the node embeddings at keyframe $t$ as $h_i^{(0)} = \mathbf{X}_i^{(t)}$.
The embeddings are then updated over $L=3$ GAT layers as follows:
\begin{equation}
\resizebox{.9\columnwidth}{!}{$
h_i^{(l)} = \sigma\Bigl( \text{Dropout}\bigl( \text{BN}\bigl( \text{GAT}^{(l)}(h_i^{(l-1)}, \mathbf{E}') \bigr) \bigr) \Bigr), \quad l=1,2,3,
$}
\end{equation}
where $\text{GAT}^{(l)}(\cdot)$ denotes the $l^\text{th}$ GAT layer employing a multi-head attention mechanism with two heads per layer, $\text{BN}(\cdot)$ is Batch Normalization, $\sigma(\cdot)$ represents the ReLU activation function, and Dropout is applied for regularization. A global pooling operation, $\text{Pool}(\cdot)$, aggregates the final node embeddings $h_i^{(3)}$ into a graph-level embedding:
\begin{equation}
g_i^{(t)} = \text{Pool}\Bigl( h_i^{(3)} \Bigr).
\end{equation}

The sequence of graph-level embeddings $\{g_i^{(t)}\}_{t \in \mathcal{K}_i}$, where $\mathcal{K}_i$ is the set of keyframe indices for demonstration $d_i$, serves as input to the temporal encoding stage. This stage employs a two-layer GRU in an autoregressive fashion. Specifically, at each time step $t$, the GRU takes the current graph-level embedding $g_i^{(t)}$ along with the previous hidden state $s_i^{(t-1)}$ (initialized as $s_i^{(0)} = \mathbf{0}$) to produce an updated hidden state:
\begin{equation}
s_i^{(t)} = \text{GRU}\Bigl( g_i^{(t)}, s_i^{(t-1)} \Bigr).
\end{equation}
The output at each time step, $s_i^{(t)}$, is then passed through a fully connected layer $f_{\text{FC}}(\cdot)$ to yield the predicted action for that time step:
\begin{equation}
\hat{\mathbf{S}}_i^{(t)} = f_{\text{FC}}\Bigl( s_i^{(t)} \Bigr).
\end{equation}

By integrating refined spatial encoding via three GAT layers—with edge connectivity informed by spectral clustering—and autoregressive temporal modeling via GRUs, our model robustly captures both inter-node relationships and their evolution over time.

\subsection{Gaussian Processes}

Each demonstration dataset \(D_k\) consists of positional coordinates for \(N\) robots and goal node at discrete timesteps, expressed as:
\begin{equation}
\resizebox{.87\columnwidth}{!}{$
D_k = \{\mathbf{x}_i^{(t)} \mid \mathbf{x}_i^{(t)} \in \mathbb{R}^2,\; i \in \{1,\dots,N,N+1\},\; t=1,\dots,T_k \},
$}
\end{equation}

where \(i\) indexes the robots (\(1,\dots,N\)) and the goal node (\(N+1\)), and \(T_k\) denotes the total number of frames in demonstration \(k\). 

Following preprocessing, complete trajectory segments were extracted from the generated temporal graph sequences. A trajectory segment \(\tau_{s\rightarrow t}^{(k)}\), corresponding to the movement of a robot from a source node \(s\) to a target node \(t\) within demonstration \(k\), is defined as:
\begin{equation}
    \tau_{s\rightarrow t}^{(k)} = \{\mathbf{x}_s^{(m)} \mid m = t_{\text{start}}, \dots, t_{\text{end}}\},
\end{equation}
where \(t_{\text{start}}\) and \(t_{\text{end}}\) represent the frame indices marking the beginning and end of a continuous robot movement segment. Each trajectory segment was annotated with its corresponding demonstration index, source-target identifiers, and temporal boundaries. This structured annotation provided a consistent dataset of trajectory segments for subsequent trajectory modeling.

To generalize demonstrations for novel contexts, GP regression encoded motion primitives for each unique source–target robot pair. Before training, trajectories were normalized, aligned to a canonical frame, and uniformly resampled into fixed-length 100-point sequences for consistency. For each aligned source-target trajectory set, two GP models (\(x\text{GP}\) and \(y\text{GP}\)) were trained to map normalized time to \(x\) and \(y\) positions, using an RBF + WhiteKernel to capture spatial correlations and noise, with training data aggregated from all resampled segments for that pair.

\begin{figure*}[ht]
\centering
\includegraphics[width=0.80\textwidth]{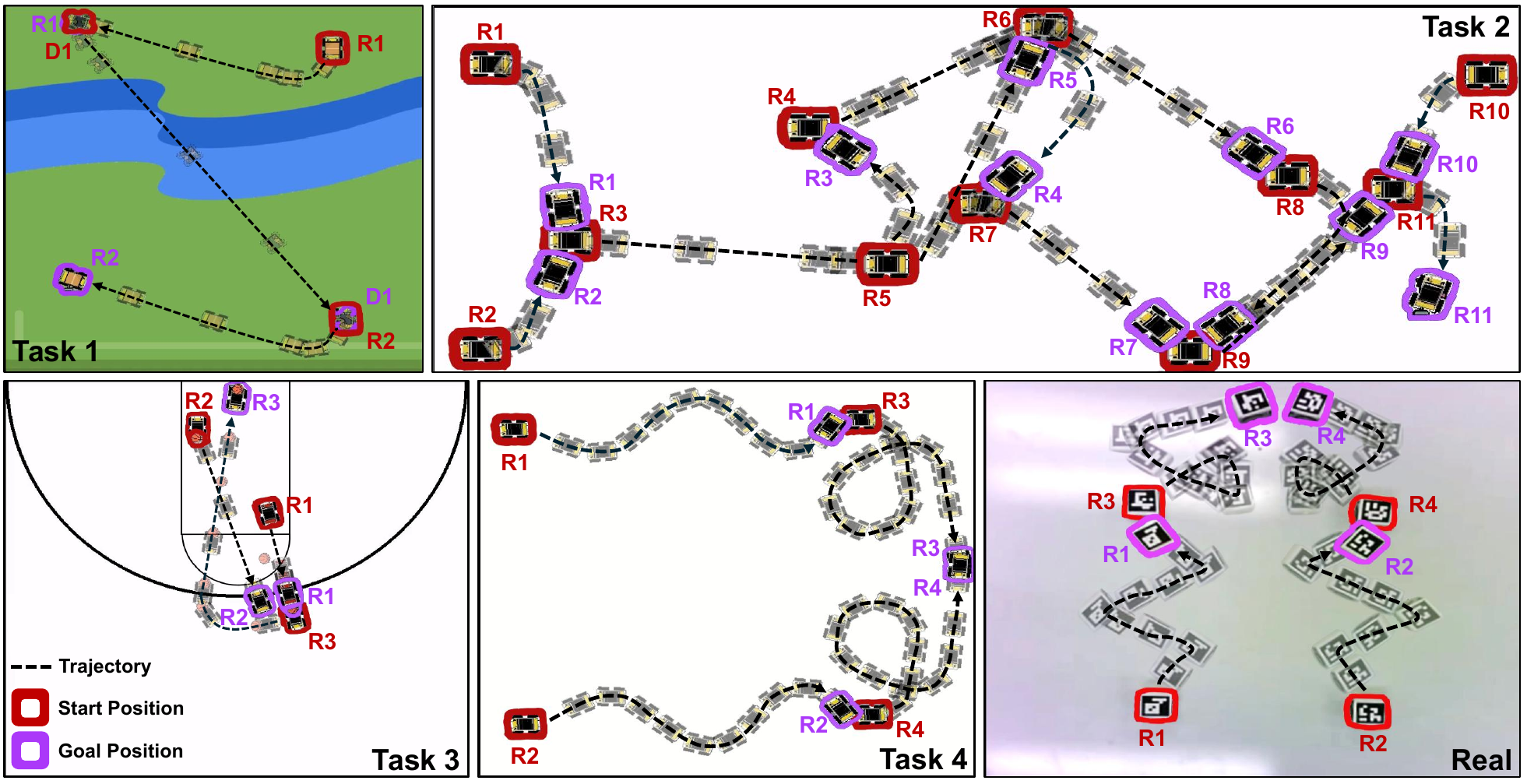}
\caption{Visualizations of experimental tasks illustrating diverse multi-robot scenarios: Task 1 features a heterogeneous team of three robots completing a five-step collaborative transport operation; Task 2 evaluates scalability, with eleven robots executing a ten-step coordinated sequence; Task 3 simulates a sports-inspired scenario involving three heterogeneous robots performing a four-step sequence of diverse actions; Task 4 focuses on spatial generalization, requiring four robots to execute complex curved and spiral trajectories in a two-step task; and the Real task reproduces Task 4 in a physical environment to validate both sequence prediction and trajectory generation under realistic conditions, as it presents temporal and spatial learning challenges.}
\label{fig_3}
\vspace{-5pt}
\end{figure*}

\subsection{Execution Phase}

At runtime, the DDACE framework operates hierarchically, with the TGN serving as the high-level policy and the GP-based motion generator as the low-level controller. The TGN infers the next high-level action step at an event-driven inference frequency: a new decision is made only after all currently active robots have completed their ongoing motion segments. This preserves the intended partial-order structure of the task and prevents premature re-planning. This design is not intended to reproduce demonstrations through frame-level trajectory replay. Instead, synchronization at completion events enforces explicit action boundaries, enabling the model to learn and reason over discrete sequence transitions.

Once the next action step is determined, the GP-based motion generator produces a motion primitive that generalizes the demonstrated trajectory for the corresponding source–target pair. Given arbitrary start and end positions \(\mathbf{x}_{\text{new}}^{(\text{start})}\) and \(\mathbf{x}_{\text{new}}^{(\text{end})}\), the trained \(x\text{GP}\) and \(y\text{GP}\) models predict a canonical trajectory \(\mathbf{x}_{GP}^{(c)}(t')\) for normalized time \(t' \in [0,1]\). This canonical form is then adapted to the new spatial context through inverse scaling, rotation, and translation:
\begin{equation}
\resizebox{.85\columnwidth}{!}{$
    \mathbf{x}_{\text{new}}(t') = R^{-1}(\theta_{\text{new}})\left(\mathbf{x}_{GP}^{(c)}(t') \cdot \|\mathbf{x}_{\text{new}}^{(\text{end})} - \mathbf{x}_{\text{new}}^{(\text{start})}\|\right) + \mathbf{x}_{\text{new}}^{(\text{start})},
$}
\end{equation}
where \(\theta_{\text{new}}\) is the orientation of the target relative to the start, computed as:
\begin{equation}
    \theta_{\text{new}} = \arctan2\left(y_{\text{new}}^{(\text{end})}-y_{\text{new}}^{(\text{start})},\; x_{\text{new}}^{(\text{end})}-x_{\text{new}}^{(\text{start})}\right).
\end{equation}

\section{Experiments and Results}
\label{experiments}
\subsection{Experimental Setup}
\label{experimental_setup}

The experimental tasks are intended as controlled task families constructed to analyze structural factors in few-shot multi-robot coordination. As summarized in Table~\ref{table1}, each task isolates specific variables, such as sequence length, team size, action-space complexity, and trajectory geometry, allowing systematic examination of how temporal–spatial decomposition and relational sparsity influence learning stability under limited demonstrations.

\begin{table}[t]
\caption{Summary of Task Properties}
\centering
\resizebox{\linewidth}{!}{\begin{tblr}{
  cells = {c},
  hline{1-2,6} = {-}{},
}
\textbf{Task Description} & \textbf{Multi Sequence} & \textbf{Heterogeneous} & \textbf{Trajectory} & \textbf{Multi Action} \\
Task 1                    &   \textcolor{green}{\checkmark}  &   \textcolor{green}{\checkmark}  &   \textcolor{red}{\ding{55}}  &   \textcolor{green}{\checkmark}  \\
Task 2                    &   \textcolor{green}{\checkmark}  &   \textcolor{red}{\ding{55}}  &   \textcolor{red}{\ding{55}}  &   \textcolor{red}{\ding{55}}  \\
Task 3                    &   \textcolor{green}{\checkmark}  &   \textcolor{green}{\checkmark}  &   \textcolor{green}{\checkmark}  &   \textcolor{green}{\checkmark}  \\
Task 4                    &   \textcolor{green}{\checkmark}  &   \textcolor{red}{\ding{55}}  &   \textcolor{green}{\checkmark}  &   \textcolor{red}{\ding{55}}  
 \vspace{-5pt}                     
\end{tblr}}
\label{table1}
\vspace{-5pt}
\end{table}

Task~1 evaluates heterogeneous multi-sequence coordination, featuring three robots with different modalities (e.g., UGV and UAV) performing a five-step collaborative transport operation. Task~2 serves as a scalability benchmark, with eleven homogeneous robots executing a ten-step sequence in which dependencies span the entire team; by excluding heterogeneity and complex trajectories, it isolates the challenge of sustaining long-horizon temporal dependencies in large-scale teams. Task~3 assesses heterogeneous multi-action coordination in a sports-inspired setting adapted from a commonly used basketball ball-screening tactic, where coordinated positioning, passing, and movement outmaneuver a defender. Three heterogeneous robots perform a four-step sequence involving passing, defensive screening, and goal approach, reflecting a widely recognized and generalizable cooperation pattern. Task~4 emphasizes spatial generalization in trajectory learning: in a two-step sequence, four robots execute symmetric curved paths followed by spiral-like trajectories toward their goals, testing the GP model’s ability to reproduce and adapt complex, non-linear paths to new spatial configurations while preserving shape fidelity.

Additional experiments were conducted to assess the influence of task complexity on learning outcomes. In Tasks~2 and~3, the number of robots and the action space per robot were systematically varied. This allowed an analysis of how these variables impact the learning outcomes of the model. Demonstration data were preprocessed and partitioned into training and testing sets, with 90\% allocated for training and 10\% for evaluation.

Qualitative assessment was performed in simulation during the execution phase, as shown in Fig.~\ref{fig_3}. Given the initial positions of robots, objects, and goals, the trained sequence predictor produced action assignments for each robot; when motion was required, corresponding trajectories were generated by the GP model. Execution sequences proceeded iteratively until task completion. Quantitative evaluation was performed against two baselines: a conventional GNN with fixed connectivity and a language-model-based (LLM) method for sequence inference. For the LLM-based baseline, the model received the task prompt and the same raw demonstrations used by DDACE, including robot/object identifiers, time-indexed coordinates, and observed action labels, and was prompted to infer the corresponding action sequence. An ablation study was also conducted by removing the spectral clustering module and substituting a fully connected edge structure. For each task, evaluation was performed over 15 trials under novel start-goal configurations, and all reported metrics are averaged across these trials.

The performance of the proposed model and baseline methods was assessed using metrics, partially adapted from the evaluation framework in~\cite{kannan2024smart}. 
In our framework, the evaluation explicitly distinguishes between temporal sequence prediction and spatial trajectory generation, which are decoupled in DDACE’s architecture:

\begin{itemize}[leftmargin=*]
    \item Temporal Component – Sequential Action Prediction:
    \begin{itemize}[leftmargin=*]
        \item \textit{Overall Success Rate (OSR):} Measures the proportion of tasks in which all robots successfully executed their entire action sequences and satisfied all task-specific goal conditions. This reflects end-to-end temporal coordination quality.
        \item \textit{Sequence Success Rate (SSR):} Measures the percentage of action sequences completed without error, regardless of whether the overall task succeeded. Even when the overall task fails, this metric provides an intuitive measure of how many individual sequences were executed successfully, reflecting per-sequence temporal accuracy.
        \item \textit{Goal Condition Recall (GCR):} Measures the fraction of required goal states achieved by each individual robot, based on its ability to reach designated goal positions or conditions. This captures per-agent temporal accuracy.
    \end{itemize}

    \item Spatial Component – Trajectory Generation via Gaussian Processes:
    \begin{itemize}[leftmargin=*]
        \item \textit{Fréchet Distance (FD):} Quantifies the geometric similarity between the GP-generated trajectories and the demonstrated ones, after normalization for scale and position.
    \end{itemize}
\end{itemize}

\begin{table*}[ht]
\caption{Quantitative performance comparison of DDACE against baseline across all experimental tasks.}
\centering
\resizebox{\textwidth}{!}{\begin{tblr}{
  cells = {c},
  cell{1}{1} = {r=2}{},
  cell{1}{2} = {c=4}{},
  cell{1}{6} = {c=4}{},
  cell{1}{10} = {c=4}{},
  cell{1}{14} = {c=4}{},
  hline{1,3,7} = {-}{},
  hline{2} = {2-4,6-8,10-12,15-16}{},
  hline{2} = {5,9,13,17}{r},
  hline{2} = {14}{l},
}
\textbf{Methods }                   & \textbf{Task 1 } &              &              &             & \textbf{Task 2} &              &              &             & \textbf{Task 3} &              &              &             & \textbf{Task 4} &              &              &             \\
                                    & \textbf{OSR}     & \textbf{SSR} & \textbf{GCR} & \textbf{FD} & \textbf{OSR}    & \textbf{SSR} & \textbf{GCR} & \textbf{FD} & \textbf{OSR}    & \textbf{SSR} & \textbf{GCR} & \textbf{FD} & \textbf{OSR}    & \textbf{SSR} & \textbf{GCR} & \textbf{FD} \\
\textbf{DDACE}        &         \textbf{1}         &       \textbf{1}       &        \textbf{1}      &     \textbf{0.02}    &  \textbf{1}    &    \textbf{1}  &    \textbf{1}      &      \textbf{0}    &   \textbf{1}  &   \textbf{1}   &   \textbf{1}  &      \textbf{0}   &    \textbf{1}   &    \textbf{1}     & \textbf{1}   &    \textbf{0.04}     \\
DDACE (without spectral clustering) &       0       &        0.2      &      0.74        &      0.02       &       0          &      0.3        &         0.59     &       0      &        0         &     0.25        &       0.50       &      0       &         0        &        0.5      &        0.74      &    0.04         \\
GNN      &        0          &       0       &       0.33       &      0.67       &        0         &      0        &         0.18     &       1      &           0      &          0    &      0.16        &       0.93      &           0      &      0       &         0.25     &       1      \\
LLM (GPT-5.2)     &         1         &        1      &        1      &       -      &        0         &      0.5        &       0.64       &      -       &        0         &        0.25      &       0.25       &        -     &          1       &       1       &         1     &       -      
\end{tblr}}
\label{table2}
\vspace{-10pt}
\end{table*}

\subsection{Qualitative Analysis}
\label{qualitative_analysis}
Execution results for all four tasks are visualized in Fig.~\ref{fig_3}. In Task~1, a heterogeneous transport scenario is depicted, where Robot R1 transfers cargo to Drone D1, which crosses a river and hands it to Robot R2. R2 then completes the delivery to the designated goal. This task involves both wheeled and aerial robots, each operating in a multi-action space including driving, stopping, and cargo transfer. The DDACE framework was observed to perform reliably under such heterogeneous coordination demands.

Task~2 involves a large-scale sequential operation executed by eleven robots. The task begins with R1 and R2 moving toward R3, followed by R3 proceeding toward R4 via R5. Subsequently, R4 and R5 move toward R6, with R5 halting upon arrival while R4 continues to R7. Independent movements of R6 and R7 to R8 and R9 follow, leading to R8 moving toward R9, then R9 and R10 converging at R11, which finally moves to the goal. This scenario was designed to test scalability, and stable execution was observed over a ten-step sequence.

Task~3 draws inspiration from a ball-screening strategy in basketball. Initially, R2 passes to teammate R3 while opponent R1 approaches for defense. R2 then screens R1, allowing R3 to drive past and proceed toward the goal. This task combines heterogeneous robot roles and a complex multi-action space, including ball passing and dribbling. Accurate trajectory generation was required to ensure R3 moved beyond the screen. The framework effectively captured both the sequential dependencies and trajectory complexity involved.

Task~4 features a simple two-step sequence but emphasizes spatial complexity. Robots R1 and R2 move toward R3 and R4 along symmetric curved paths, after which R3 and R4 navigate spiral-like trajectories to reach the goal. This task was used to assess the model’s ability to learn and reproduce nontrivial path geometries. The results confirmed that DDACE could successfully generalize and replicate such intricate trajectories.

\subsection{Quantitative Analysis}
\label{quantitative_analysis}
A quantitative evaluation of the proposed framework was performed across all four tasks using the metrics defined in Section~\ref{experimental_setup}. As shown in Table~\ref{table2}, DDACE consistently achieved the highest performance, obtaining perfect scores in OSR, SSR, and GCR across all tasks. Low FD values ranging from 0.02 in Task~1 to 0.04 in Task~4 indicate that the generated trajectories closely matched those from the demonstrations, demonstrating accurate spatial reproduction and sequence fidelity.

Several baseline comparisons also serve as ablation-like tests of DDACE design choices. Replacing the refined edge index from the spectral clustering stage with a fully connected graph structure removes the targeted graph refinement step in the temporal module. This configuration, while functioning as a baseline, led to significant drops in OSR and SSR, highlighting the importance of identifying critical inter-robot dependencies prior to sequence learning. Without spectral clustering, the lack of structural priors caused poorer coordination and disrupted temporal alignment. Although GCR remained relatively higher, this did not consistently translate to successful task completion due to disrupted temporal alignment.

Similarly, the end-to-end GNN baseline can be viewed as a proxy ablation of DDACE’s temporal–spatial decoupling, since it jointly learns action sequences and trajectories in a single model. This configuration demonstrated poor performance, with zero OSR and SSR, low GCR, and FD values near 1.0, suggesting that jointly modeling temporal coordination and spatial motion increases hypothesis coupling and limits few-shot generalization. In contrast, DDACE’s modular design preserves coherent sequence prediction and high-fidelity trajectory generation under limited demonstrations.

The LLM baseline was used to interpret demonstration data and generate task sequences. It performed well in Tasks~1 and~4, where the sequences were largely governed by positional relationships. However, it struggled in Task~3 with heterogeneous roles and contextual action dependencies, and in Task~2 with long-horizon sequence completion. These results show that the LLM baseline can infer simple sequences, but without explicit trajectory modeling or temporal–spatial decomposition, it is currently limited in long-horizon or heterogeneous coordination tasks.

\begin{figure}
\centering
\includegraphics[width=1\columnwidth]{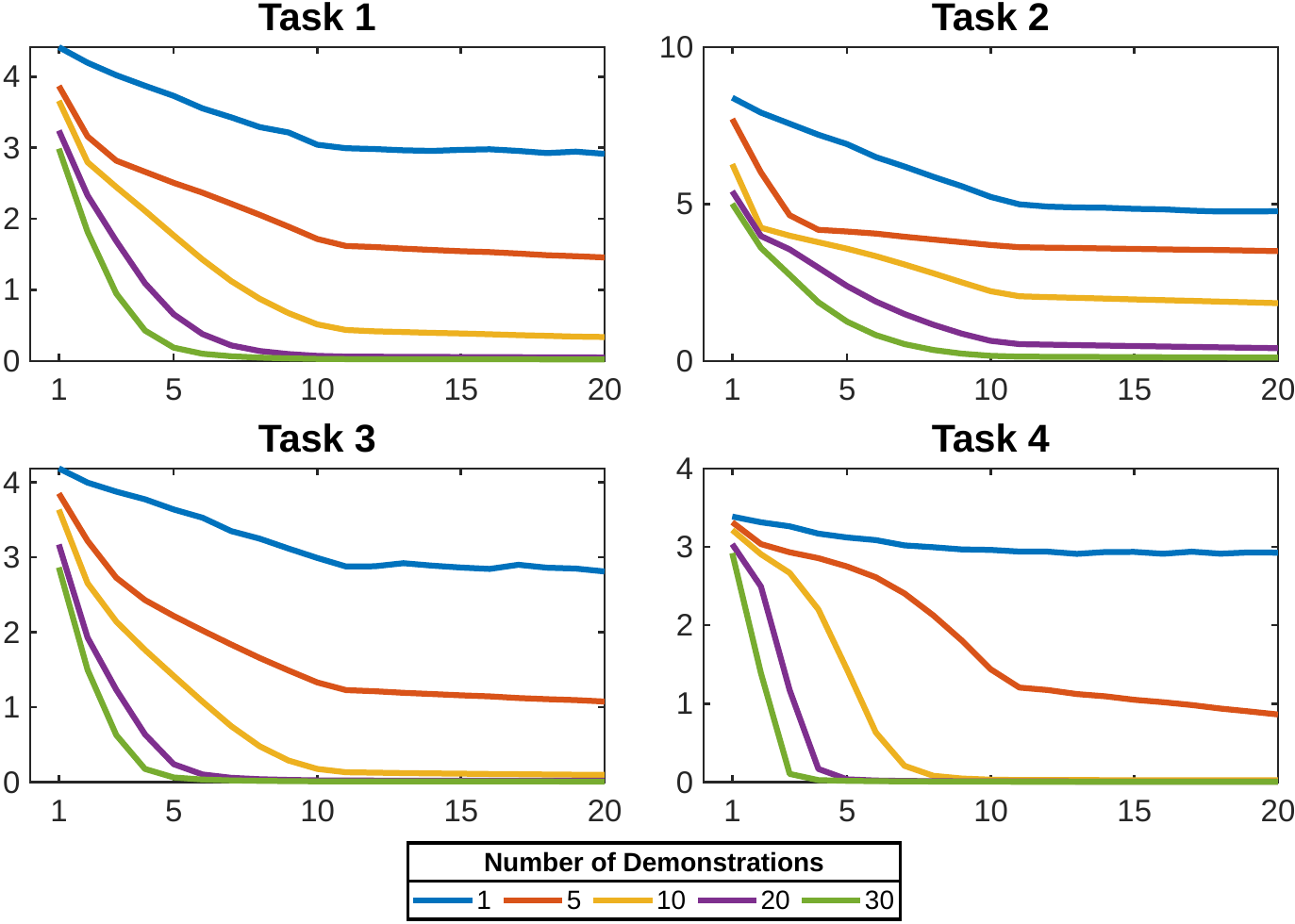}
\caption{Training loss versus epoch for Tasks~1-4.  Each subplot shows the evolution of the training loss (y-axis) over epochs (x-axis) for different numbers of demonstrations. }
\label{fig_4}
\vspace{-13pt}
\end{figure}

To analyze the effect of demonstration quantity on performance, Fig.~\ref{fig_4} presents training loss curves for Tasks~1–4. Across all tasks, additional demonstrations led to faster convergence and lower final loss. Tasks~1 and~4, with simpler sequence structures, converged reliably with as few as 10 demonstrations. Task~2, which involves long-horizon dependencies, required more data for stable learning, while Task~3, with heterogeneous roles and multi-action sequences, also benefited from additional demonstrations. These results show that DDACE remains effective under severe few-shot constraints, though additional data improves stability for more complex tasks. While this evaluation assumes optimal, noise-free demonstrations, imperfect demonstrations may slow convergence or bias learned outputs. However, DDACE’s decoupling and relational sparsification may mitigate mild noise by emphasizing recurring coordination patterns.

\subsubsection{Case Study 1: Effect of Number of Robots}
This case study examines how the number of robots affects learning dynamics. Fig.~\ref{fig:cs}-left shows training loss trends for Task~2 with 4 to 11 robots using 30 demonstrations per setting. Smaller teams converged faster and reached lower final loss due to reduced coordination complexity. In the 4-node case, minimal interdependencies enabled near-zero loss rapidly. Configurations with 6 to 7 nodes preserved the core task structure and maintained efficient learning, whereas 9--11 node scenarios showed slower convergence and higher final loss indicating that increased team size introduced coordination challenges that complicated the learning process. These results emphasize the scalability limits in few-shot settings, where larger team sizes amplify structural complexity and increase learning difficulty, highlighting the type of coordination challenge that DDACE’s temporal-spatial decoupling is designed to mitigate.

\subsubsection{Case Study 2: Impact of Action Space Complexity}
\begin{figure}[t]
\centering
\includegraphics[width=0.90\columnwidth]{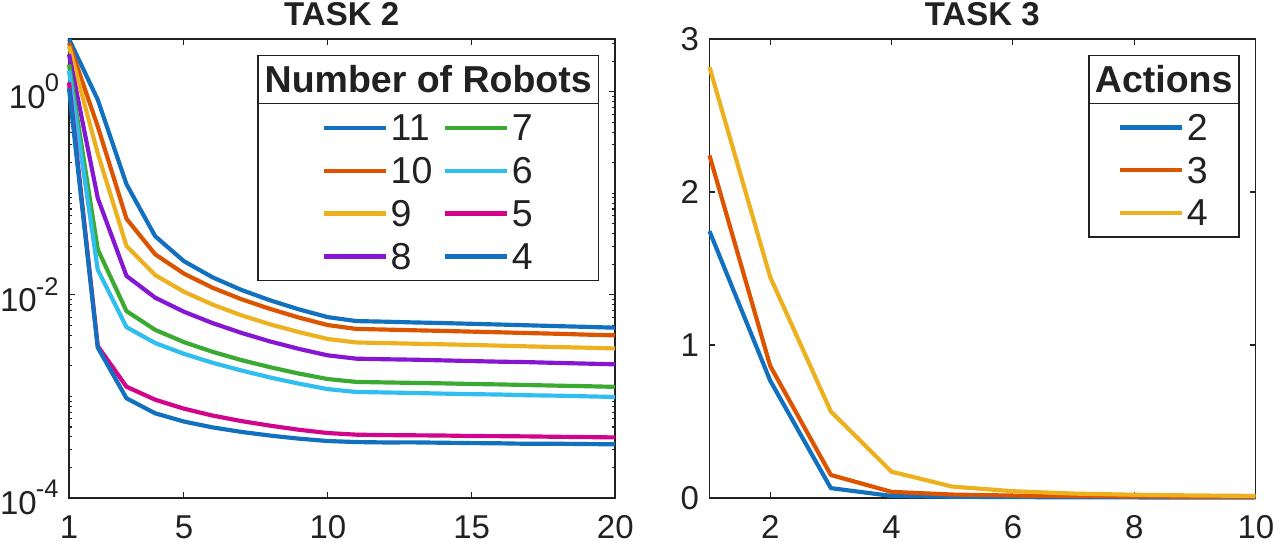}
\vspace{-5pt}
\caption{Training loss analysis across varying robot configurations and action spaces. (\textbf{Left}) Log‐scaled training loss versus epoch for different robot team sizes. (\textbf{Right}) Training loss versus epoch for varying action‐space complexities.}

\label{fig:cs}
\vspace{-13pt}
\end{figure}

Similarly, this case study examines the effect of action space size on convergence behavior. The influence of action space size was examined in Task~3 using three configurations-2, 3, and 4 actions-with 30 demonstrations each (Fig.~\ref{fig:cs}-right). Simpler action spaces (move, stop) enabled rapid and stable convergence. The introduction of a third action (dribble) increased temporal dependencies and slightly delayed convergence. The 4-action setting (including pass) further increased the learning burden, resulting in higher initial loss and slower convergence. These findings indicate that increasing action diversity elevates the reasoning complexity, which in turn slows training, particularly in tasks involving rich inter-agent dependencies or decision-making under temporal constraints. Nonetheless, the model eventually attained low loss in all cases, owing to DDACE’s decoupled framework, which divides key functionalities into separate components, enabling successful learning even under minimal dataset conditions.

\subsection{Real-World Deployment}
\label{real_world}
To evaluate real-world applicability, DDACE was deployed on physical robots to reproduce Task~4 from simulation. This task was chosen because it combines temporal and spatial learning challenges, with sequential structures and complex trajectories (e.g., curves and spirals), making it well-suited for validating both the sequence predictor and trajectory generator under realistic conditions.

Experiments were conducted on a tabletop setup using Hamster mobile robots~\cite{lee2021investigating} with Bluetooth control and overhead ArUco-based vision tracking~\cite{GARRIDOJURADO20142280}. Models were transferred from simulation without fine-tuning, and as shown in Fig.~\ref{fig_3} and the supplementary video, planned behaviors and trajectories were reproduced with high fidelity, demonstrating DDACE’s few-shot generalization to physical environments.

\section{Conclusion}
This paper presented DDACE as a structural approach to few-shot multi-robot coordination, examining temporal-spatial decomposition and relational sparsification as inductive biases for stabilizing learning under limited demonstrations. By factorizing high-level action sequencing from geometric trajectory synthesis, the framework constrains hypothesis complexity while preserving sequence-level reasoning and geometry-consistent motion generation. The proposed architecture is particularly effective for stage-based coordination tasks with reusable motion primitives and partial-order temporal structure. Future work may extend this decomposition toward more tightly interleaved and highly reactive scenarios requiring finer-grained asynchronous coordination. Additional directions include improving robustness to imperfect demonstrations through retrieval-augmented or semi-supervised few-shot learning~\cite{RSSfew,wu2025semi}, and evaluating structural inductive bias design in larger-scale real-world datasets such as SportVU~\cite{sportvu}.


\bibliographystyle{IEEEtran}
\bibliography{references}

@inproceedings{kannan2024smart,
  title={Smart-llm: Smart multi-agent robot task planning using large language models},
  author={Kannan, Shyam Sundar and Venkatesh, Vishnunandan LN and Min, Byung-Cheol},
  booktitle={2024 IEEE/RSJ International Conference on Intelligent Robots and Systems (IROS)},
  pages={12140--12147},
  year={2024},
  organization={IEEE}
}

@software{sportvu,
  author = {Linou, Kostya and Linou, Dzmitryi and Boer, Martijn de},
  month = {9},
  title = {{NBA-Player-Movements}},
  url = {https://github.com/linouk23/NBA-Player-Movements},
  year = {2016}
}

@article{lee2021investigating,
  title={Investigating the effect of deictic movements of a multi-robot},
  author={Lee, Ahreum and Jo, Wonse and Kannan, Shyam Sundar and Min, Byung-Cheol},
  journal={International Journal of Human--Computer Interaction},
  volume={37},
  number={3},
  pages={197--210},
  year={2021},
  publisher={Taylor \& Francis}
}

@inproceedings{wu2024multi,
  title={A Multi-Robot Task Allocation Method Based on Graph Attention Network and Unsupervised Learning},
  author={Wu, Zirui and Li, Zhen and Zhu, Dong and Liao, Qiuhan and Yao, Weiran},
  booktitle={2024 IEEE International Conference on Unmanned Systems (ICUS)},
  pages={1222--1227},
  year={2024},
  organization={IEEE}
}

@inproceedings{chen2021spatio,
  title={Spatio-Temporal Graph Policy Gradients for Multi-Robot Formation Control},
  author={Chen, Shaofeng and Cao, Yang and Kang, Yu and Di, Jian and Sun, BingYu and Wang, Xuefeng},
  booktitle={2021 7th International Conference on Big Data and Information Analytics (BigDIA)},
  pages={436--439},
  year={2021},
  organization={IEEE}
}

@inproceedings{goeckner2024graph,
  title={Graph neural network-based multi-agent reinforcement learning for resilient distributed coordination of multi-robot systems},
  author={Goeckner, Anthony and Sui, Yueyuan and Martinet, Nicolas and Li, Xinliang and Zhu, Qi},
  booktitle={2024 IEEE/RSJ International Conference on Intelligent Robots and Systems (IROS)},
  pages={5732--5739},
  year={2024},
  organization={IEEE}
}

@article{zhang2022h2gnn,
  title={H2GNN: Hierarchical-hops graph neural networks for multi-robot exploration in unknown environments},
  author={Zhang, Hao and Cheng, Jiyu and Zhang, Lin and Li, Yibin and Zhang, Wei},
  journal={IEEE Robotics and Automation Letters},
  volume={7},
  number={2},
  pages={3435--3442},
  year={2022},
  publisher={IEEE}
}

@article{goarin2024graph,
  title={Graph neural network for decentralized multi-robot goal assignment},
  author={Goarin, Manohari and Loianno, Giuseppe},
  journal={IEEE Robotics and Automation Letters},
  year={2024},
  publisher={IEEE}
}

@inproceedings{blumenkamp2022framework,
  title={A framework for real-world multi-robot systems running decentralized GNN-based policies},
  author={Blumenkamp, Jan and Morad, Steven and Gielis, Jennifer and Li, Qingbiao and Prorok, Amanda},
  booktitle={2022 International Conference on Robotics and Automation (ICRA)},
  pages={8772--8778},
  year={2022},
  organization={IEEE}
}

@inproceedings{chekakta2024towards,
  title={Towards learning-based distributed task allocation approach for multi-robot system},
  author={Chekakta, Zakaria and Aouf, Nabil and Govindaraj, Shashank and Polisano, Fabio and De Cubber, Geert},
  booktitle={2024 10th International Conference on Automation, Robotics and Applications (ICARA)},
  pages={34--39},
  year={2024},
  organization={IEEE}
}

@inproceedings{li2020graph,
  title={Graph neural networks for decentralized multi-robot path planning},
  author={Li, Qingbiao and Gama, Fernando and Ribeiro, Alejandro and Prorok, Amanda},
  booktitle={2020 IEEE/RSJ international conference on intelligent robots and systems (IROS)},
  pages={11785--11792},
  year={2020},
  organization={IEEE}
}

@inproceedings{chandrasekaran2023survey,
  title={A Survey of Path Planning and Navigation in Multi-Robotic Systems},
  author={Chandrasekaran, Balasubramaniyan},
  booktitle={2023 1st International Conference on Advanced Engineering and Technologies (ICONNIC)},
  pages={38--42},
  year={2023},
  organization={IEEE}
}

@inproceedings{venkatesh2024learning,
  title={Learning from Demonstration Framework for Multi-Robot Systems Using Interaction Keypoints and Soft Actor-Critic Methods},
  author={Venkatesh, Vishnunandan LN and Min, Byung-Cheol},
  booktitle={2024 IEEE/RSJ International Conference on Intelligent Robots and Systems (IROS)},
  pages={10754--10761},
  year={2024},
  organization={IEEE}
}

@article{argall2009survey,
  title={A survey of robot learning from demonstration},
  author={Argall, Brenna D and Chernova, Sonia and Veloso, Manuela and Browning, Brett},
  journal={Robotics and autonomous systems},
  volume={57},
  number={5},
  pages={469--483},
  year={2009},
  publisher={Elsevier}
}

@article{ravichandar2020recent,
  title={Recent advances in robot learning from demonstration},
  author={Ravichandar, Harish and Polydoros, Athanasios S and Chernova, Sonia and Billard, Aude},
  journal={Annual review of control, robotics, and autonomous systems},
  volume={3},
  pages={297--330},
  year={2020},
  publisher={Annual Reviews}
}

@article{billard2016learning,
  title={Learning from humans},
  author={Billard, Aude G and Calinon, Sylvain and Dillmann, R{\"u}diger},
  journal={Springer handbook of robotics},
  pages={1995--2014},
  year={2016},
  publisher={Springer}
}

@inproceedings{huang2017vision,
  title={A vision-guided multi-robot cooperation framework for learning-by-demonstration and task reproduction},
  author={Huang, Bidan and Ye, Menglong and Lee, Su-Lin and Yang, Guang-Zhong},
  booktitle={2017 IEEE/RSJ International Conference on Intelligent Robots and Systems (IROS)},
  pages={4797--4804},
  year={2017},
  organization={IEEE}
}

@article{burgard2005coordinated,
  title={Coordinated multi-robot exploration},
  author={Burgard, Wolfram and Moors, Mark and Stachniss, Cyrill and Schneider, Frank E},
  journal={IEEE Transactions on robotics},
  volume={21},
  number={3},
  pages={376--386},
  year={2005},
  publisher={IEEE}
}

@inproceedings{darmanin2017review,
  title={A review on multi-robot systems categorised by application domain},
  author={Darmanin, Rachael N and Bugeja, Marvin K},
  booktitle={2017 25th mediterranean conference on control and automation (MED)},
  pages={701--706},
  year={2017},
  organization={IEEE}
}

@article{chernova2009interactive,
  title={Interactive policy learning through confidence-based autonomy},
  author={Chernova, Sonia and Veloso, Manuela},
  journal={Journal of Artificial Intelligence Research},
  volume={34},
  pages={1--25},
  year={2009}
}

@inproceedings{knepper2013ikeabot,
  title={Ikeabot: An autonomous multi-robot coordinated furniture assembly system},
  author={Knepper, Ross A and Layton, Todd and Romanishin, John and Rus, Daniela},
  booktitle={2013 IEEE International conference on robotics and automation},
  pages={855--862},
  year={2013},
  organization={IEEE}
}

@inproceedings{martins2010learning,
  title={Learning multirobot joint action plans from simultaneous task execution demonstrations.},
  author={Martins, Murilo Fernandes and Demiris, Yiannis},
  booktitle={AAMAS},
  pages={931--938},
  year={2010}
}

@inproceedings{freelan2014towards,
  title={Towards rapid multi-robot learning from demonstration at the robocup competition},
  author={Freelan, David and Wicke, Drew and Sullivan, Keith and Luke, Sean},
  booktitle={Robot Soccer World Cup},
  pages={369--382},
  year={2014},
  organization={Springer}
}

@inproceedings{sullivan2011hierarchical,
  title={Hierarchical multi-robot learning from demonstration},
  author={Sullivan, Keith and Luke, Sean},
  booktitle={Proceedings of the Robotics: Science and Systems Conference},
  year={2011}
}

@inproceedings{balakuntala2021learning,
  title={Learning Multimodal Contact-Rich Skills from Demonstrations Without Reward Engineering},
  author={Balakuntala, Mythra V and Kaur, Upinder and Ma, Xin and Wachs, Juan and Voyles, Richard M},
  booktitle={2021 IEEE International Conference on Robotics and Automation (ICRA)},
  pages={4679--4685},
  year={2021},
  organization={IEEE}
}

@article{brunke2022safe,
  title={Safe learning in robotics: From learning-based control to safe reinforcement learning},
  author={Brunke, Lukas and Greeff, Melissa and Hall, Adam W and Yuan, Zhaocong and Zhou, Siqi and Panerati, Jacopo and Schoellig, Angela P},
  journal={Annual Review of Control, Robotics, and Autonomous Systems},
  volume={5},
  pages={411--444},
  year={2022},
  publisher={Annual Reviews}
}

@article{simoes2020dataset,
  title={A dataset schema for cooperative learning from demonstration in multi-robot systems},
  author={Sim{\~o}es, Marco AC and da Silva, Robson Marinho and Nogueira, Tatiane},
  journal={Journal of Intelligent \& Robotic Systems},
  volume={99},
  number={3},
  pages={589--608},
  year={2020},
  publisher={Springer}
}

@inproceedings{freelan2015towards,
  title={Towards rapid multi-robot learning from demonstration at the robocup competition},
  author={Freelan, David and Wicke, Drew and Sullivan, Keith and Luke, Sean},
  booktitle={RoboCup 2014: Robot World Cup XVIII 18},
  pages={369--382},
  year={2015},
  organization={Springer}
}

@article{zhu2024madiff,
  title={Madiff: Offline multi-agent learning with diffusion models},
  author={Zhu, Zhengbang and Liu, Minghuan and Mao, Liyuan and Kang, Bingyi and Xu, Minkai and Yu, Yong and Ermon, Stefano and Zhang, Weinan},
  journal={Advances in Neural Information Processing Systems},
  volume={37},
  pages={4177--4206},
  year={2024}
}

@article{meng2023offline,
  title={Offline pre-trained multi-agent decision transformer},
  author={Meng, Linghui and Wen, Muning and Le, Chenyang and Li, Xiyun and Xing, Dengpeng and Zhang, Weinan and Wen, Ying and Zhang, Haifeng and Wang, Jun and Yang, Yaodong and others},
  journal={Machine Intelligence Research},
  volume={20},
  number={2},
  pages={233--248},
  year={2023},
  publisher={Springer}
}

@article{GARRIDOJURADO20142280,
title = {Automatic generation and detection of highly reliable fiducial markers under occlusion},
journal = {Pattern Recognition},
volume = {47},
number = {6},
pages = {2280-2292},
year = {2014},
issn = {0031-3203},
author = {S. Garrido-Jurado and R. Muñoz-Salinas and F.J. Madrid-Cuevas and M.J. Marín-Jiménez}
}

@INPROCEEDINGS{RSSfew, 
    AUTHOR    = {Maximilian Du AND Suraj Nair AND Dorsa Sadigh AND Chelsea Finn}, 
    TITLE     = {{Behavior Retrieval: Few-Shot Imitation Learning by Querying Unlabeled Datasets}}, 
    BOOKTITLE = {Proceedings of Robotics: Science and Systems}, 
    YEAR      = {2023}, 
    ADDRESS   = {Daegu, Republic of Korea}, 
    MONTH     = {July}, 
    DOI       = {10.15607/RSS.2023.XIX.011} 
}

@article{wu2025semi,
    title={Semi-Supervised One Shot Imitation Learning},
    author={Wu, Philipp and Hakhamaneshi, Kourosh and Du, Yuqing and Mordatch, Igor and Rajeswaran, Aravind and Abbeel, Pieter},
    journal={Reinforcement Learning Journal},
    volume={5},
    pages={2284--2297},
    year={2025}
}

\end{document}